\documentclass[letterpaper, 10 pt, conference]{ieeeconf}  
\IEEEoverridecommandlockouts                              

\title{\bf Beating humans in a penny-matching game by leveraging \\ cognitive hierarchy theory and Bayesian learning}

 \author{Ran Tian, Nan Li, Ilya Kolmanovsky, and Anouck Girard
 \thanks{This research has been supported by the National Science Foundation Award Number CNS 1544844.}
  \thanks{Ran Tian, Nan Li, Ilya Kolmanovsky, and Anouck Girard are with the Department of Aerospace Engineering, University of Michigan, Ann Arbor, MI 48109, USA. {\tt\small \{tianran, nanli, ilya, anouck\}@umich.edu}.
     }
  }
\usepackage{tabularx}
\usepackage{float}
\usepackage[caption=false]{subfig}
\usepackage{epsfig}
\usepackage{epstopdf}
\usepackage{bbm}
\usepackage{epsfig} 
\usepackage{amsmath} 
\usepackage{amssymb}  
\usepackage{mathrsfs}
\usepackage{bm}
\usepackage{epstopdf}
\usepackage{color}
\usepackage{etoolbox}

\usepackage{comment}
\usepackage{diagbox}
\usepackage[ruled, lined, longend, linesnumbered]{algorithm2e}
\usepackage{booktabs,multirow}
\usepackage{bigstrut}
\usepackage{cite}
\usepackage{graphicx}

\usepackage{balance}
\begin{document}
\maketitle

\begin{abstract}
It is a long-standing goal of artificial intelligence (AI) to be superior to human beings in decision making. Games are suitable for testing AI capabilities of making good decisions in non-numerical tasks. In this paper, we develop a new AI algorithm to play the penny-matching game considered in Shannon's ``mind-reading machine'' (1953) against human players. In particular, we exploit cognitive hierarchy theory and Bayesian learning techniques to continually evolve a model for predicting human player decisions, and let the AI player make decisions according to the model predictions to pursue the best chance of winning. Experimental results show that our AI algorithm beats $27$ out of $30$ volunteer human players.
\end{abstract}

\section{Introduction}

Developing artificial intelligence (AI) to beat humans in strategic games has been drawing attention/interest of researchers for decades \cite{shannon1950chess,campbell2002deep,silver2016mastering,silver2017mastering,brown2018superhuman,shannon1953mind,hagelbarger1956seer,ali2000playing,pozzato2013towards,zink2019predictive}. Although AI algorithms targeted at specific games may not directly contribute to solving practical engineering problems other than those in the gambling industry, the theories developed alongside can be used to attack many problems of similar natures and of greater significance \cite{shannon1950chess}. 

Many strategic games, for example, the games of chess, Go, and poker, require the players to do intensive calculations to identify winning strategies. AI algorithms for these games often rely on the super computing power of modern computers to beat human players \cite{campbell2002deep,silver2016mastering,silver2017mastering,brown2018superhuman}. In many other strategic games, computing power may have a less decisive effect on the game result. For instance, this holds for the cases where perfectly rational decision strategies are well-known, including the game of matching pennies (or ``odds and evens'') \cite{mookherjee1994learning} and the game of rock-paper-scissors \cite{wang2014social}. 
For such games, recognizing the decision pattern of the opponent human player often plays a vital role in developing winning strategies for the AI player \cite{shannon1953mind,hagelbarger1956seer,ali2000playing,pozzato2013towards,zink2019predictive}.

In this paper, we focus on the game of matching pennies. We develop an AI to play the game repeatedly against a human player, during which the AI decision strategy is continually evolved to pursue the best chance of winning.

Such a problem has been considered by C. E. Shannon in his seminal paper \cite{shannon1953mind}, where he named his AI a ``mind-reading machine,'' followed by  D. W. Hagelbarger in \cite{hagelbarger1956seer}, where the AI won $5,218$ times out of $9,795$ plays. The principle behind their AI algorithms is the hypothesis that human players are not able to generate {\it i.i.d.}\footnote{Independent and identically distributed.} random numbers but tend to follow certain patterns depending on the previous results to make their new decisions. Their AI algorithms pursue the identification of these patterns and assume that the human player will follow the same patterns the next time the same situation arises. Some later works, including \cite{ali2000playing,pozzato2013towards,zink2019predictive} for the game of rock-paper-scissors, essentially follow the same principle to design their AI algorithms.

In this paper, we propose a new AI algorithm, which leverages cognitive hierarchy theory \cite{stahl1995players,costa2006cognition,camerer2004cognitive} and Bayesian learning. We hypothesize that human players follow certain ``patterns'' in making decisions, but differently from \cite{shannon1953mind} and \cite{hagelbarger1956seer}, we explicitly characterize these ``patterns'' based on the human player's ``level of reasoning,'' introduced in cognitive hierarchy theory. Furthermore, we assume that humans follow these patterns probabilistically, and we use Bayesian learning techniques to identify the associated probabilities.

In summary, the contributions of this paper are:

1) We develop a new AI algorithm to play the penny-matching game, which beats $90\%$ of volunteer human players in our experiments.

2) Our algorithm exploits the level-k framework \cite{stahl1995players,costa2006cognition} of cognitive hierarchy theory, which has not been considered in previously developed AI algorithms for penny-matching or rock-paper-scissors games \cite{shannon1953mind,hagelbarger1956seer,ali2000playing,pozzato2013towards,zink2019predictive}. 

3) Although cognitive hierarchy theory has been exploited for modeling human-agent behavior in some other application domains, such as automotive \cite{7993050,li2019decision}, aerospace \cite{yildiz2013predicting}, and cyber-physical security  \cite{kanellopoulos2019non} applications, probabilistic reasoning-level transitions in sequential decision-making scenarios, considered in this paper, have not been incorporated in these previous works. And the results of this paper suggest that such reasoning-level transitions exist in human decision making and can be modeled.

4) In the light of 3), we envision that the general approach to modeling human behavior proposed in this paper can find its utility in a broader range of application scenarios involving human-machine interactions, such as the scenarios in \cite{7993050,li2019decision,yildiz2013predicting,kanellopoulos2019non}.

\section{Mathematical formulation of the penny-matching game}

The penny-matching game under consideration, denoted by $\mathcal{G}$, is a two-player zero-sum game with the normal-form representation given in Table~\ref{tab:game}. Formally, let $u^i$ denote a decision of player $i$, $i \in \mathcal{P} = \{1,2\}$, taking values in the set $\mathcal{U} = \{0,1\}$. The payoffs of the two players, $(r^1,r^2)$, as functions of $(u^1,u^2)$ are defined as follows:
\begin{subequations}\label{equ:payoffs}
\begin{align}
r^1 &= 1-2\, \text{mod}(u^1 + u^2,2), \\
r^2 &= -r^1.
\end{align}

\end{subequations}
\begin{table}[h!]
\centering
\begin{tabular}{|c|c|c|}
\hline
\diagbox{Player 1}{Player 2} & $0$ & $1$ \\\hline
$0$ & $(1,-1)$ & $(-1,1)$ \\\hline
$1$ & $(-1,1)$ & $(1,-1)$ \\\hline
\end{tabular}
\caption{Game in normal-form representation.}
\label{tab:game}
\vspace{-0.2in}
\end{table}

We let a human player, as player $1$, and an artificial intelligence (AI), as player $2$, play the game $\mathcal{G}$ repeatedly. For convenience, we use the subscript $t \in \mathbb{N} \cup \{0\}$ to denote the round of the game. For instance, $u^1_t$ denotes the human player's decision for game round $t$, and $r^1_t$ denotes her obtained payoff in this round.

If $r^i_t>0$, then we say that player $i$ wins the round $t$. It is easily seen from \eqref{equ:payoffs} that there is always one and only one of the two players winning a round. Our goal is to design a strategy for the AI so that it has a higher winning rate than the human player, where the winning rate is defined as the number of wins divided by the total number of game rounds.

It is clear from \eqref{equ:payoffs} that if the human player's decision $u^1_t$ can be correctly predicted, the AI can win the round $t$ by using the following decision strategy:
\begin{equation}
    u^2_t  = 1 - \hat{u}^1_t,
\end{equation}
where $\hat{u}^1_t$ denotes the predicted value of $u^1_t$.

Therefore, the goal to win can be achieved through developing a model of the human player that can predict $u^1_t$ with high accuracy. 

It is well-known that the above repeated game has a unique Nash equilibrium, which involves independent repetition of the stage-game equilibrium strategy, i.e., for each player to play according to an {\it i.i.d} process where at each stage $0$ or $1$ is chosen with equal probability $0.5$. We call this the Nash-equilibrium strategy for $\mathcal{G}$. In particular, as long as one of the two players applies this Nash-equilibrium strategy, the game will end in a draw in expectation.

\section{Modeling human player based on cognitive hierarchy theory}\label{sec: human model}

\subsection{Level-k models of the human player}
\label{sec: human level-k model}

Cognitive hierarchy theory (CHT) characterizes human decision-making processes based on assumptions of bounded rationality and iterated reasoning. In the level-k framework of CHT, a human decision-maker is assumed to make decisions based on a finite number of reasoning steps, called ``level." In the setting of single-shot games, in particular, a level-$k$ player assumes that the other player(s) are level-$(k-1)$, predicts their decisions based on this assumption, and makes her own decision as the optimal response to the predicted decisions of the other players \cite{stahl1995players,costa2006cognition}. 

In order to formulate the level-$k$, $k = 0,1,\dots$, decision strategies of the two players in our game $\mathcal{G}$, we start from defining the level-$0$ decision rules of the two players as follows:
\begin{subequations}\label{equ:level_0}
\begin{align}
    \hat{u}^{1,0}_t  &= u^2_{t-1}, \\
    \hat{u}^{2,0}_t  &= 1 - u^1_{t-1}.
\end{align}
\end{subequations}
The above level-$0$ decision rules are based on the ``naive'' thought that the other player will make the same decision as in the previous round, which may represent a player's instinctive response to the game. 

On the basis of the level-$0$ decision rules \eqref{equ:level_0}, the level-$k$ decision rules of the two players, with $k \ge 1$, are as follows:
\begin{subequations}\label{equ:level_k_1}
\begin{align}
    \hat{u}^{1,k}_t  &= \hat{u}^{2,k-1}_t, \\
    \hat{u}^{2,k}_t  &= 1 - \hat{u}^{1,k-1}_t,
\end{align}
\end{subequations}
i.e., the level-$k$ decision of player $i$ optimally responds to the level-($k-1$) decision of player ($3-i$) in terms of maximizing player $i$'s own payoff.

For a given pair $(u^1_{t-1},u^2_{t-1}) \in \mathcal{U} \times \mathcal{U}$, the level-$k$ decisions of the two players for $k = 0,1,2,\dots$, are summarized in Table~\ref{tab:level_k}, where we use $[\kappa]$ to denote the set of non-negative integers $\kappa'$ that satisfy $\text{mod}(\kappa',4)=\kappa$.

We let $\sigma_t \in \mathbb{N} \cup \{0\}$ denote the human player's level of reasoning for round $t$. In principle, if $\sigma_t \in [\kappa]$ for some $\kappa = 0,1,2$ or $3$, then $u^1_t$ is determined as $u^1_t = \hat{u}^{1,\sigma_t}_t = \hat{u}^{1,[\kappa]}_t$, which can be read from Table~\ref{tab:level_k}. However, to account for the sub-optimality and variability in human decision making, we assume that if $\sigma_t \in [\kappa]$, then the human player makes decisions according to
\begin{subequations}\label{equ:level_k}
\begin{align}
    & \mathbb{P}(u^1_t = \hat{u}^{1,\sigma_t}_t) = \frac{e^{\theta}}{e^{\theta}+e^{-\theta}}, \\
    & \mathbb{P}(u^1_t = 1 - \hat{u}^{1,\sigma_t}_t) = \frac{e^{-\theta}}{e^{\theta}+e^{-\theta}},
\end{align}
\end{subequations}
which is based on the ``softmax'' decision rule \cite{sutton2018reinforcement} with $\theta>0$ being a tuning parameter. We remark that the ``softmax'' decision rule is a typical choice for modeling human decisions \cite{morgenstern1953theory}.

\begin{table}[h]
\centering
\begin{tabular}{|c|c|c|}
\hline
\diagbox{$k$}{Player} & $1$ & $2$ \\\hline
$[0]$ & $u^2_{t-1}$ & $1- u^1_{t-1}$ \\\hline
$[1]$ & $1- u^1_{t-1}$ & $1-u^2_{t-1}$ \\\hline
$[2]$ & $1-u^2_{t-1}$ & $u^1_{t-1}$ \\\hline
$[3]$ & $u^1_{t-1}$ & $u^2_{t-1}$ \\\hline
\end{tabular}
\caption{Level-$k$ decisions of players $1$ and $2$.}
\label{tab:level_k}
\vspace{-0.36in}
\end{table}

\subsection{Transitions of human player's reasoning level}
\label{sec: level transition}

In a single-shot game, a player has only one chance to determine her reasoning level, relying on which to make her decision. In a repeated game, in contrast, a player can adjust her level in each round, for instance, according to whether she is winning or losing.

We assume that the human player in our repeated game $\mathcal{G}$ will probabilistically adjust her reasoning level in each round according to the game result of the previous round, in particular, based on the following transition model,
\begin{subequations}\label{equ:level_transition}
\begin{align}
& \mathbb{P}(\sigma_t \in [i] \,|\, \sigma_{t-1} \in [j], r^1_{t-1} = 1) = p^+_{(i+1),(j+1)}, \\
& \mathbb{P}(\sigma_t \in [i] \,|\, \sigma_{t-1} \in [j], r^1_{t-1} = -1) = p^-_{(i+1),(j+1)},
\end{align}
\end{subequations}
defined for all $i,j \in \{0,1,2,3\}$, where $\mathbb{P}(\cdot|\cdot)$ represents conditional probabilities.

However, due to the fact that different humans may have different transition models, the transition matrices $p^+, p^- \in \{p \in [0,1]^{4 \times 4} \,|\, \sum_{i =1}^{4} p_{i,j} = 1, j = 1,2,3,4\}$ are not a priori known, but have to be estimated during the game. Note that there are $12$ unknown parameters for each of $p^+$ and $p^-$, which poses a requirement of a large set of data for their estimation.

Therefore, we pursue a simplification of the transition model \eqref{equ:level_transition} by leveraging the following two observations:

1) If the human player won the previous round, i.e., $u^{1}_{t-1} = u^2_{t-1}$, then $\hat{u}^{1,[0]}_t = \hat{u}^{1,[3]}_t$ and $\hat{u}^{1,[1]}_t = \hat{u}^{1,[2]}_t$.

2) If the human player lost the previous round, i.e., $u^{1}_{t-1} = 1 - u^2_{t-1}$, then $\hat{u}^{1,[0]}_t = \hat{u}^{1,[1]}_t$ and $\hat{u}^{1,[2]}_t = \hat{u}^{1,[3]}_t$.

In other words, to predict the human player's action for the next round, there is no need to distinguish her level between $[0]$ and $[3]$, and between $[1]$ and $[2]$ if she won the previous round. And similarly, there is no need to distinguish her level between $[0]$ and $[1]$, and between $[2]$ and $[3]$ if she lost the previous round. 

On the basis of the above observations, we consider a simplified transition model as follows: 
\begin{subequations}\label{equ:level_transition_2}
\begin{align}
& \mathbb{P}(\sigma_t \in \Sigma_1^+ \,|\, \sigma_{t-1} \in \Sigma_1^+, r^1_{t-1} = 1) = q_1^+, \\
& \mathbb{P}(\sigma_t \in \Sigma_2^+ \,|\, \sigma_{t-1} \in \Sigma_2^+, r^1_{t-1} = 1) = q_2^+, \\
& \mathbb{P}(\sigma_t \in \Sigma_1^- \,|\, \sigma_{t-1} \in \Sigma_1^-, r^1_{t-1} = -1) = q_1^-, \\
& \mathbb{P}(\sigma_t \in \Sigma_2^- \,|\, \sigma_{t-1} \in \Sigma_2^-, r^1_{t-1} = -1) = q_2^-,
\end{align}
\end{subequations}
where $\Sigma_1^+ = \{[0],[3]\}$, $\Sigma_2^+ = \{[1],[2]\}$, $\Sigma_1^- = \{[0],[1]\}$, and $\Sigma_2^- = \{[2],[3]\}$. Note that the probabilistic transitions from $\Sigma_1^+$ to $\Sigma_2^+$ and from $\Sigma_2^+$ to $\Sigma_1^+$ under $r^1_{t-1} = 1$ as well as those from $\Sigma_1^-$ to $\Sigma_2^-$ and from $\Sigma_2^-$ to $\Sigma_1^-$ under $r^1_{t-1} = -1$ can be computed based on the law of total probability. For instance, $\mathbb{P}(\sigma_t \in \Sigma_2^+ \,|\, \sigma_{t-1} \in \Sigma_1^+, r^1_{t-1} = 1) = 1 - q_1^+$. Furthermore, we assume that the probability of the event $\sigma_t \in \Sigma_i^{\pm}$ conditioned on $\sigma_{t-1} \in \Sigma_j^{\pm}$ and $r^1_{t-1} = \pm 1$ is independent of all other events for every pair of $i,j \in \{0,1\}$.

Suppose that $\sigma_{t-1}$, as well as $q_1^+$, $q_2^+$, $q_1^-$ and $q_2^-$, is known. Then, depending on $r^1_{t-1} = 1$ or $-1$, the probabilities of set membership $\sigma_t \in \Sigma_1^+$ and $\sigma_t \in \Sigma_2^+$, or the probabilities of $\sigma_t \in \Sigma_1^-$ and $\sigma_t \in \Sigma_2^-$ can be computed, based on which $\hat{u}^{1,\sigma_t}_t$ can be probabilistically predicted. Indeed, when the exact values of $(\sigma_{t-1},q_1^+,q_2^+,q_1^-,q_2^-)$ are not known, a distribution on $\{[0],[1],[2],[3]\} \times [0,1]^4$ characterizing the probability of $(\sigma_{t-1},q_1^+,q_2^+,q_1^-,q_2^-)$ taking each value of $\{[0],[1],[2],[3]\} \times [0,1]^4$ is sufficient for the above computation and prediction. Specifically, let $\pi_{t-1}$ denote such a probability distribution, then
\small
\begin{align}
 \mathbb{P}(\sigma_t \in \Sigma_1^+) &= \int_{\Sigma_1^+ \times [0,1]^4} q_1^+ \text{d}\pi_{t-1} + \int_{\Sigma_{2}^+ \times [0,1]^4} (1-q_{2}^+)\, \text{d} \pi_{t-1},
\end{align}
\normalsize
if $r^1_{t-1} = 1$, and 
\small
\begin{align}
\mathbb{P}(\sigma_t \in \Sigma_1^-) &= \int_{\Sigma_1^- \times [0,1]^4} q_1^- \text{d}\pi_{t-1} + \int_{\Sigma_{2}^- \times [0,1]^4} (1-q_{2}^-)\, \text{d} \pi_{t-1},
\end{align}
\normalsize
if $r^1_{t-1} = -1$.

To facilitate numerical implementation, we assume that $q_i^{\pm}$ takes values in a finite set $Q \subset [0,1]$, where $Q$ can be a grid on $[0,1]$. In this case, the probability distribution $\pi_{t-1}$ is discrete, and the formula becomes
\small
\begin{align}
\label{equ:P1_plus}
& \mathbb{P}(\sigma_t \in \Sigma_1^+) = \sum_{q_1^+ \in Q} q_1^+ \Big(\sum_{\sigma \in \Sigma_1^+} \sum_{q_2^+ \in Q} \sum_{q_1^- \in Q} \sum_{q_2^- \in Q} \nonumber \\ & \pi_{t-1}(\sigma,q_1^+,q_2^+,q_1^-,q_2^-)\Big) + \sum_{q_2^+ \in Q} (1-q_2^+) \Big( \nonumber\\[-2pt]
& \sum_{\sigma \in \Sigma_2^+} \sum_{q_1^+ \in Q} \sum_{q_1^- \in Q} \sum_{q_2^- \in Q} \pi_{t-1}(\sigma,q_1^+,q_2^+,q_1^-,q_2^-)\Big),
\end{align}
\normalsize
if $r^1_{t-1} = 1$, and 
\small
\begin{align}
\label{equ:P1_minus}
& \mathbb{P}(\sigma_t \in \Sigma_1^-) = \sum_{q_1^- \in Q} q_1^- \Big(\sum_{\sigma \in \Sigma_1^-} \sum_{q_2^- \in Q} \sum_{q_1^+ \in Q} \sum_{q_2^+ \in Q} \nonumber \\ & \pi_{t-1}(\sigma,q_1^+,q_2^+,q_1^-,q_2^-)\Big) + \sum_{q_2^- \in Q} (1-q_2^-) \Big(\nonumber \\[-2pt]
& \sum_{\sigma \in \Sigma_2^-} \sum_{q_1^- \in Q} \sum_{q_1^+ \in Q} \sum_{q_2^+ \in Q} \pi_{t-1}(\sigma,q_1^+,q_2^+,q_1^-,q_2^-)\Big),
\end{align}
\normalsize
if $r^1_{t-1}=-1$.

\subsection{Bayesian learning of human player's model}
\label{sec: learning transition model}

On the basis of Sections~\ref{sec: human level-k model} and \ref{sec: level transition}, the human player's behavior is modeled based on two parts: her reasoning level $\sigma_t$ for each round $t$ and the parameters $(q_1^+,q_2^+,q_1^-,q_2^-)$ characterizing her reasoning level transitions. Unfortunately, these variables/parameters are neither a priori known nor directly observable. What can be observed are the human player's decision for each round, $u_t^1$, and the game result for each round, $r_t^1$. Note that given $(u_t^1,r_t^1)$, the knowledge of $u_t^2$ and $r_t^2$ is redundant since they can be computed using $(u_t^1,r_t^1)$ and \eqref{equ:payoffs}. 

We will use Bayesian learning techniques to learn $x_t = (\sigma_t,q_1^+,q_2^+,q_1^-,q_2^-)$ from the observable data $\xi_t = \{u^1_0,\dots,u^1_{t},r^1_0,\dots,r^1_t\}$. Specifically, we pursue a probability distribution on $\{[0],[1],[2],[3]\} \times Q^4$, characterizing our belief  about the value of $x_t$, i.e., the $\pi_{t}$ defined at the end of Section \ref{sec: level transition}, conditioned on the available data $\xi_t$.

To achieve this, we rely on a hidden Markov chain formulation and its corresponding recursive Bayesian inference formula as follows:

If $r^1_t = 1$, then we have
\small
\begin{align}\label{equ:pi_plus}    
& \pi_{t}(\Sigma_1^+ \times \{q_1^+,q_2^+,q_1^-,q_2^-\}) = \nonumber \\
& \frac{\mathbb{P}(u^1_{t}|\sigma_t \in \Sigma_1^+)\, \Pi_{t-1}^+}{\sum_{\hat{\bf q} \in Q^4} \big(\mathbb{P}(u^1_{t}|\sigma_t \in \Sigma_1^+)\,\hat{\Pi}_{t-1,1}^+ + \mathbb{P}(u^1_{t}|\sigma_t \in \Sigma_2^+)\,\hat{\Pi}_{t-1,2}^+\big)}
\end{align}
\normalsize
where $\sum_{\hat{\bf q} \in Q^4} = \sum_{\hat{q}_1^+ \in Q}\sum_{\hat{q}_2^+ \in Q} \sum_{\hat{q}_1^- \in Q} \sum_{\hat{q}_2^- \in Q}$, and
\small
\begin{subequations}\label{equ:Pi_plus}
\begin{align}
    &\Pi_{t-1}^+ = q_1^+ \pi_{t-1}(\Sigma_1^+ \times \{q_1^+,q_2^+,q_1^-,q_2^-\}) \nonumber \\
    &\quad\quad\quad + (1-q_2^+) \pi_{t-1}(\Sigma_2^+ \times \{q_1^+,q_2^+,q_1^-,q_2^-\}), \\[4pt]
    &\hat{\Pi}_{t-1,1}^+ = \hat{q}_1^+ \pi_{t-1}(\Sigma_1^+ \times \{\hat{q}_1^+,\hat{q}_2^+,\hat{q}_1^-,\hat{q}_2^-\}) \nonumber \\
    &\quad\quad\quad + (1-\hat{q}_2^+) \pi_{t-1}(\Sigma_2^+ \times \{\hat{q}_1^+,\hat{q}_2^+,\hat{q}_1^-,\hat{q}_2^-\}), \\[4pt]
    &\hat{\Pi}_{t-1,2}^+ = (1-\hat{q}_1^+) \pi_{t-1}(\Sigma_1^+ \times \{\hat{q}_1^+,\hat{q}_2^+,\hat{q}_1^-,\hat{q}_2^-\}) \nonumber \\
    &\quad\quad\quad\,\, + \hat{q}_2^+ \pi_{t-1}(\Sigma_2^+ \times \{\hat{q}_1^+,\hat{q}_2^+,\hat{q}_1^-,\hat{q}_2^-\}),
\end{align}
\end{subequations}
\normalsize
and based on \eqref{equ:level_k},
\begin{align}
& \mathbb{P}(u^1_{t}|\sigma_t \in \Sigma_1^+) = \begin{cases} \frac{e^{\theta}}{e^{\theta}+e^{-\theta}} & \text{if }  u^1_{t} = \hat{u}^{1,[0]}_t (= \hat{u}^{1,[3]}_t), \\
\frac{e^{-\theta}}{e^{\theta}+e^{-\theta}} & \text{if }  u^1_{t} = \hat{u}^{1,[1]}_t (= \hat{u}^{1,[2]}_t), 
\end{cases} \nonumber \\[3pt]
& \mathbb{P}(u^1_{t}|\sigma_t \in \Sigma_2^+) = \begin{cases} \frac{e^{-\theta}}{e^{\theta}+e^{-\theta}} & \text{if }  u^1_{t} = \hat{u}^{1,[0]}_t (= \hat{u}^{1,[3]}_t), \\
\frac{e^{\theta}}{e^{\theta}+e^{-\theta}} & \text{if }  u^1_{t} = \hat{u}^{1,[1]}_t (= \hat{u}^{1,[2]}_t).
\end{cases}
\end{align}

Similarly, if $r^1_t = -1$, then we have
\small
\begin{align}\label{equ:pi_minus}
& \pi_{t}(\Sigma_1^- \times \{q_1^+,q_2^+,q_1^-,q_2^-\}) = \nonumber \\
& \frac{\mathbb{P}(u^1_{t}|\sigma_t \in \Sigma_1^-)\, \Pi_{t-1}^-}{\sum_{\hat{\bf q} \in Q^4} \mathbb{P}(u^1_{t}|\sigma_t \in \Sigma_1^-)\,\hat{\Pi}_{t-1,1}^- + \mathbb{P}(u^1_{t}|\sigma_t \in \Sigma_2^-)\,\hat{\Pi}_{t-1,2}^- }
\end{align}
\normalsize
where 
\small
\begin{subequations}\label{equ:Pi_minus}
\begin{align}
    &\Pi_{t-1}^- = q_1^- \pi_{t-1}(\Sigma_1^- \times \{q_1^+,q_2^+,q_1^-,q_2^-\}) \nonumber \\
    &\quad\quad\quad + (1-q_2^-) \pi_{t-1}(\Sigma_2^- \times \{q_1^+,q_2^+,q_1^-,q_2^-\}), \\[4pt]
    &\hat{\Pi}_{t-1,1}^- = \hat{q}_1^- \pi_{t-1}(\Sigma_1^- \times \{\hat{q}_1^+,\hat{q}_2^+,\hat{q}_1^-,\hat{q}_2^-\}) \nonumber \\
    &\quad\quad\quad + (1-\hat{q}_2^-) \pi_{t-1}(\Sigma_2^- \times \{\hat{q}_1^+,\hat{q}_2^+,\hat{q}_1^-,\hat{q}_2^-\}), \\[4pt]
    &\hat{\Pi}_{t-1,2}^- = (1-\hat{q}_1^-) \pi_{t-1}(\Sigma_1^- \times \{\hat{q}_1^+,\hat{q}_2^+,\hat{q}_1^-,\hat{q}_2^-\}) \nonumber \\
    &\quad\quad\quad\,\, + \hat{q}_2^- \pi_{t-1}(\Sigma_2^- \times \{\hat{q}_1^+,\hat{q}_2^+,\hat{q}_1^-,\hat{q}_2^-\}),
\end{align}
\end{subequations}
\normalsize
and based on \eqref{equ:level_k},
\begin{align}
& \mathbb{P}(u^1_{t}|\sigma_t \in \Sigma_1^-) = \begin{cases} \frac{e^{\theta}}{e^{\theta}+e^{-\theta}} & \text{if }  u^1_{t} = \hat{u}^{1,[0]}_t (= \hat{u}^{1,[1]}_t), \\
\frac{e^{-\theta}}{e^{\theta}+e^{-\theta}} & \text{if }  u^1_{t} = \hat{u}^{1,[2]}_t (= \hat{u}^{1,[3]}_t),
\end{cases} \nonumber \\[3pt]
& \mathbb{P}(u^1_{t}|\sigma_t \in \Sigma_2^-) = \begin{cases} \frac{e^{-\theta}}{e^{\theta}+e^{-\theta}} & \text{if }  u^1_{t} = \hat{u}^{1,[0]}_t (= \hat{u}^{1,[1]}_t), \\
\frac{e^{\theta}}{e^{\theta}+e^{-\theta}} & \text{if }  u^1_{t} = \hat{u}^{1,[2]}_t (= \hat{u}^{1,[3]}_t).
\end{cases}
\end{align}

Note that in computing \eqref{equ:Pi_plus} or \eqref{equ:Pi_minus}, we need to use $\pi_{t-1}$, the belief distribution of $x_{t-1}$ on $\{[0],[1],[2],[3]\} \times Q^4$ conditioned on the available data $\xi_{t-1}$. However, \eqref{equ:pi_plus} or \eqref{equ:pi_minus} only provides us with partial information of $\pi_{t}$, that is, $\pi_{t}(\Sigma_1^+ \times \{q_1^+,q_2^+,q_1^-,q_2^-\})$ or $\pi_{t}(\Sigma_1^- \times \{q_1^+,q_2^+,q_1^-,q_2^-\})$. Note that $\pi_{t}(\Sigma_2^+ \times \{q_1^+,q_2^+,q_1^-,q_2^-\}) = 1 - \pi_{t}(\Sigma_1^+ \times \{q_1^+,q_2^+,q_1^-,q_2^-\})$ and $\pi_{t}(\Sigma_2^- \times \{q_1^+,q_2^+,q_1^-,q_2^-\}) = 1- \pi_{t}(\Sigma_1^- \times \{q_1^+,q_2^+,q_1^-,q_2^-\})$. 

To make the propagation \eqref{equ:pi_plus} or \eqref{equ:pi_minus} (which one is used depends on the game result $r^1_t$) recursively computable for all $t$, we need to reconstruct the distribution $\pi_{t}$ from the partial information $\pi_{t}(\Sigma_1^+ \times \{q_1^+,q_2^+,q_1^-,q_2^-\})$ or $\pi_{t}(\Sigma_1^- \times \{q_1^+,q_2^+,q_1^-,q_2^-\})$. To do so, we rely on the following assumptions:
\small
\begin{align}\label{equ:indistinguishable}
& \pi_{t}([i] \times \{q_1^+,q_2^+,q_1^-,q_2^-\}):\pi_{t}([j] \times \{q_1^+,q_2^+,q_1^-,q_2^-\}) = \nonumber \\
& \pi_{t-1}([i] \times \{q_1^+,q_2^+,q_1^-,q_2^-\}):\pi_{t-1}([j] \times \{q_1^+,q_2^+,q_1^-,q_2^-\})
\end{align}
\normalsize
holds for the pairs $(i,j) = ([0],[3])$ and $(i,j) = ([1],[2])$ if $r^1_t = 1$, and holds for the pairs $(i,j) = ([0],[1])$ and $(i,j) = ([2],[3])$ if $r^1_t = -1$, and for all $\{q_1^+,q_2^+,q_1^-,q_2^-\} \in Q^4$, meaning that our relative degree of belief in any two indistinguishable\footnote{In terms of corresponding to identical $\hat{u}^{1,\sigma_t}_t$.} levels follows its previous value.

On the basis of \eqref{equ:pi_plus}, \eqref{equ:pi_minus}, and \eqref{equ:indistinguishable}, $\pi_{t}$ can be computed using $\pi_{t-1}$, $u_t^1$, and $r_t^1$ for all $t$.

\section{Decision strategy for the AI player}\label{sec: AI decision}

Using the algorithm \eqref{equ:pi_plus}-\eqref{equ:indistinguishable}, after each round $t-1$, we can obtain a belief distribution $\pi_{t-1}$ characterizing the human player's model. Then, we compute $\mathbb{P}(\sigma_t \in \Sigma_1^+)$ and $\mathbb{P}(\sigma_t \in \Sigma_2^+) = 1-\mathbb{P}(\sigma_t \in \Sigma_1^+)$ using \eqref{equ:P1_plus} if $r^1_{t-1}=1$, or compute $\mathbb{P}(\sigma_t \in \Sigma_1^-)$ and $\mathbb{P}(\sigma_t \in \Sigma_2^-) = 1-\mathbb{P}(\sigma_t \in \Sigma_1^-)$ using \eqref{equ:P1_minus} if $r^1_{t-1}=-1$.

Suppose that $\sigma_t$ is known. Then, we let the AI mimic a human player's decision strategy, i.e., a ``softmax'' decision rule similar to \eqref{equ:level_k} as follows:
\begin{subequations}\label{equ:AI_pure}
\begin{align}
    & \mathbb{P}(u^2_t = 1-\hat{u}^{1,\sigma_t}_t|\sigma_t) = \frac{e^{\theta}}{e^{\theta}+e^{-\theta}}, \\
    & \mathbb{P}(u^2_t = \hat{u}^{1,\sigma_t}_t|\sigma_t) = \frac{e^{-\theta}}{e^{\theta}+e^{-\theta}},
\end{align}
\end{subequations}
with $\theta>0$ being the same parameter as in \eqref{equ:level_k}. We remark that although the AI does not need to mimic the sub-optimality in human decision making, the strategy \eqref{equ:AI_pure} creates some randomness in AI decisions, making it harder for the human player to identify the decision algorithm behind the AI, while guaranteeing that the probability of winning is higher than $0.5$.

Since $\sigma_t$ is not exactly known, we let the AI make decisions relying on the predicted distribution of $\sigma_t$ as follows:

If $r_{t-1}^1 = 1$, then $\hat{u}^{1,[0]}_t = \hat{u}^{1,[3]}_t = u^2_{t-1}$, and we let
\small
\begin{align}\label{equ:AI_decision_1}
    & \mathbb{P}(u^2_t = u^2_{t-1}) \nonumber \\
    &= \sum_{\sigma_t \in \{[0],[1],[2],[3]\}} \mathbb{P}(u^2_t = u^2_{t-1}|\sigma_t)\mathbb{P}(\sigma_t) \nonumber \\
    &= \sum_{\sigma_t \in \{[0],[3]\}} \frac{e^{-\theta}}{e^{\theta}+e^{-\theta}} \mathbb{P}(\sigma_t) + \sum_{\sigma_t \in \{[1],[2]\}} \frac{e^{\theta}}{e^{\theta}+e^{-\theta}} \mathbb{P}(\sigma_t) \nonumber \\
    &= \frac{e^{-\theta}}{e^{\theta}+e^{-\theta}} \mathbb{P}(\sigma_t \in \Sigma_1^+) + \frac{e^{\theta}}{e^{\theta}+e^{-\theta}} \mathbb{P}(\sigma_t \in \Sigma_2^+) \nonumber \\[2pt]
    &= \frac{e^{\theta}}{e^{\theta}+e^{-\theta}} - \frac{e^{\theta} - e^{-\theta}}{e^{\theta}+e^{-\theta}}\, \mathbb{P}(\sigma_t \in \Sigma_1^+).
\end{align}
\normalsize

Similarly, if $r_{t-1}^1 = -1$, then $\hat{u}^{1,[0]}_t = \hat{u}^{1,[1]}_t = u^2_{t-1}$, and we let 
\small
\begin{equation}\label{equ:AI_decision_2}
    \mathbb{P}(u^2_t = u^2_{t-1}) = \frac{e^{\theta}}{e^{\theta}+e^{-\theta}} - \frac{e^{\theta} - e^{-\theta}}{e^{\theta}+e^{-\theta}}\, \mathbb{P}(\sigma_t \in \Sigma_1^-).
\end{equation}
\normalsize

In turn, $\mathbb{P}(u^2_t = 1 - u^2_{t-1}) = 1 - \mathbb{P}(u^2_t = u^2_{t-1})$.

\section{Results}

\subsection{Game GUI}

\begin{figure}[h]
\begin{center}
\begin{picture}(200.0, 148)
\put(  -20,  -5){\epsfig{file=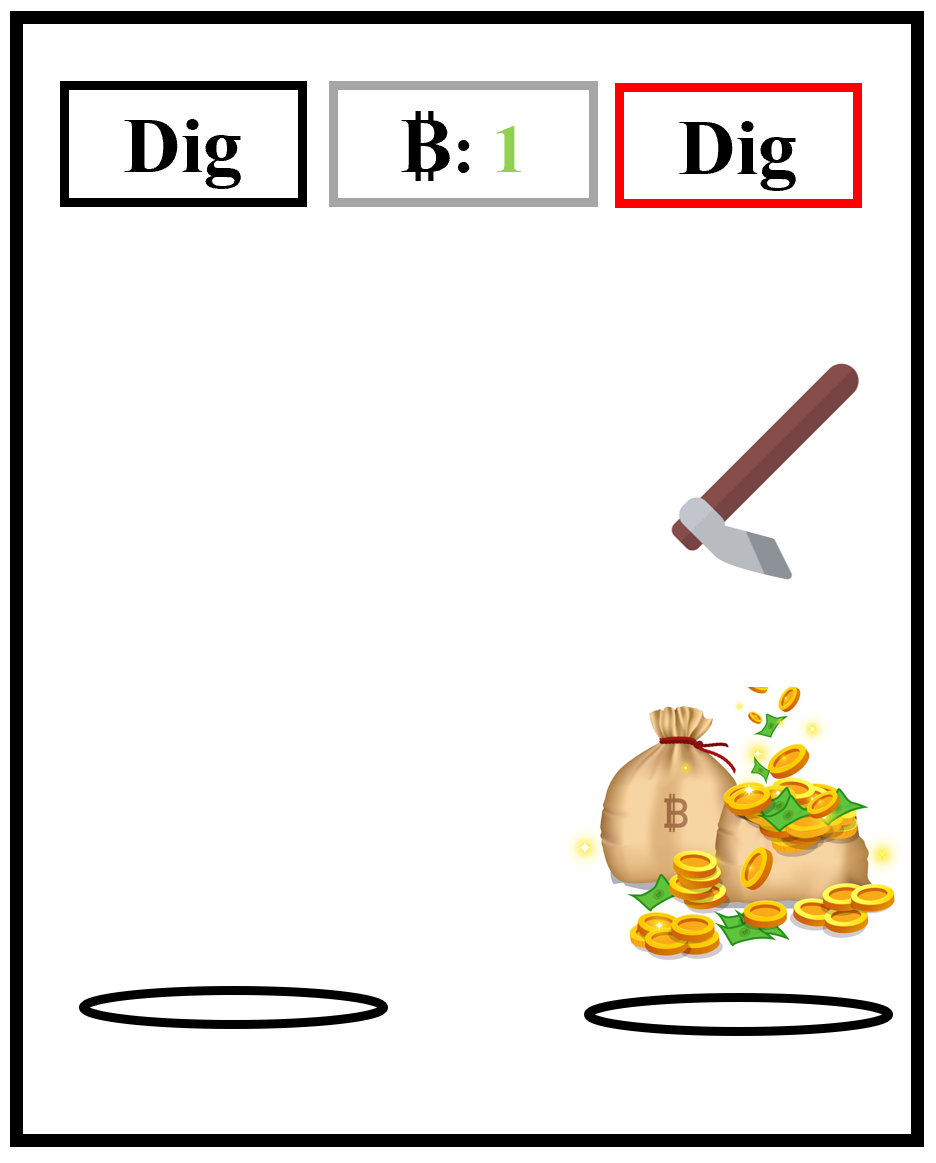,width = 0.48 \linewidth, trim=0.0cm 0.0cm 0cm 0cm,clip}}  
\put(  100,  -4.7){\epsfig{file=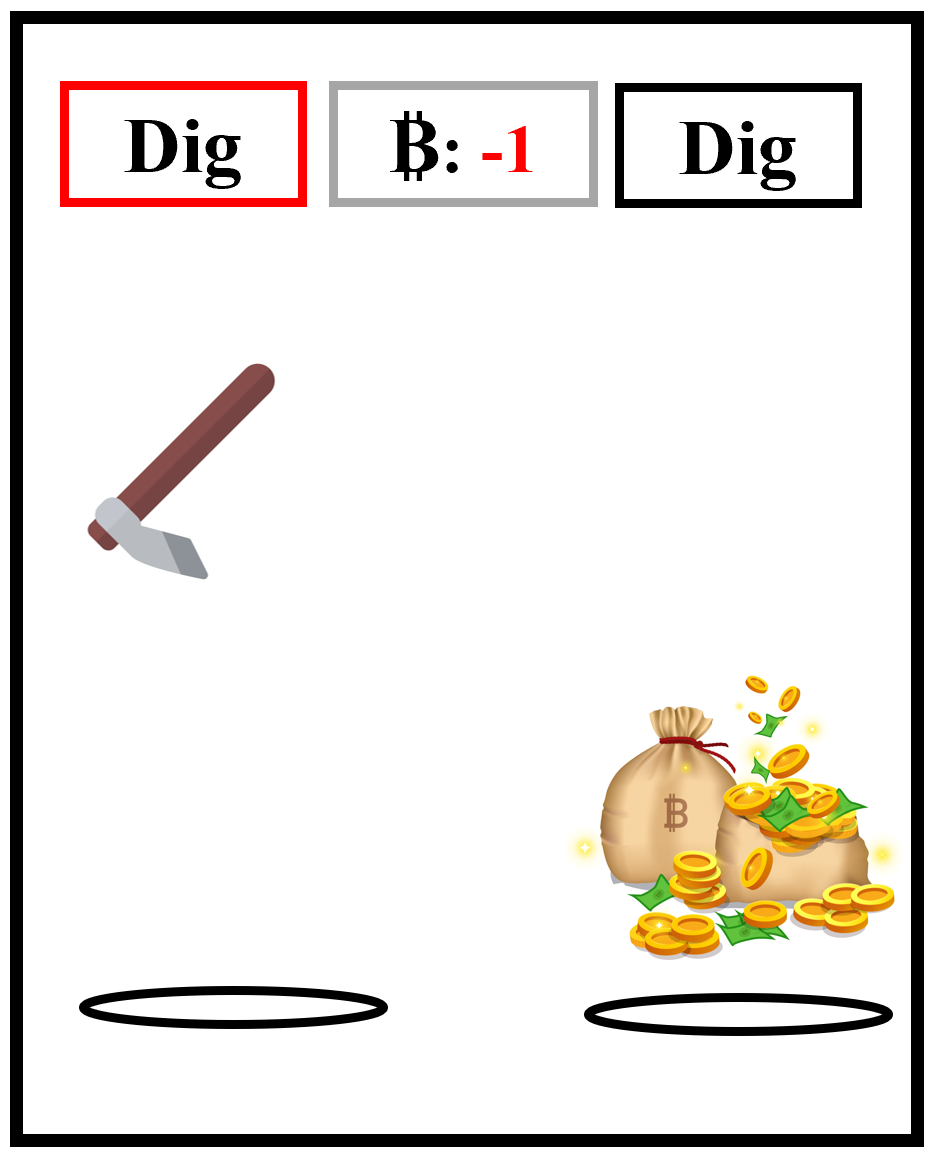,width = 0.48 \linewidth, trim=0.0cm 0.0cm 0cm 0cm,clip}}  
\small
\put(75,143){(a)}
\put(195,143){(b)}
\normalsize
\end{picture}
\end{center}
      \caption{Game GUI. (a) The human player and the AI player both choose the right side, and thus the human player wins; (b) The human player chooses the left side and the AI player chooses the right side, and thus the human player loses.}
      \label{fig: game_visual}
      \vspace{-0.12in}
\end{figure}

We design a Graphic User Interface (GUI), shown in Fig.~\ref{fig: game_visual}, to represent the game $\mathcal{G}$. In each round, the human player makes a decision between left or right to dig and the AI player makes a decision between left or right to hide the treasure. The human player gains one virtual coin ($r^1_{t}=1$) if both players choose the same side, and loses one ($r^1_{t}=-1$) otherwise. The decisions of the two players for the current round are displayed once both decisions have been made and until the human player has made her decision for the next round. The accumulated payoff of the human player up to the current round $t$, i.e., $\sum_{k=1}^{t} r^1_{k}$, is shown in the top-middle.

\subsection{Results}

We recruited human volunteers to play the game against the AI player. In particular, we let each human participant play the game twice, each time with $150$ rounds. In one of the two game-runs, the AI uses our proposed strategy leveraging cognitive hierarchy theory and Bayesian learning, described in Sections~\ref{sec: human model} and \ref{sec: AI decision}. In the algorithm \eqref{equ:pi_plus}-\eqref{equ:AI_decision_2}, we use the parameters $Q = \{0.1,0.3,0.5,0.7,0.9\}$ and $\theta = 1.5$. In the other game-run, the AI uses a Nash-equilibrium strategy, i.e., randomly chooses between left or right with equal probability in each round. The order of these two strategies used in the two game-runs is randomly determined and the human participant knows neither the decision algorithms behind the AI, nor the fact that the AI uses different strategies in the two game-runs.

\begin{figure}[ht!]
\begin{center}
\begin{picture}(200, 140)
\put(  0,  -8){\epsfig{file=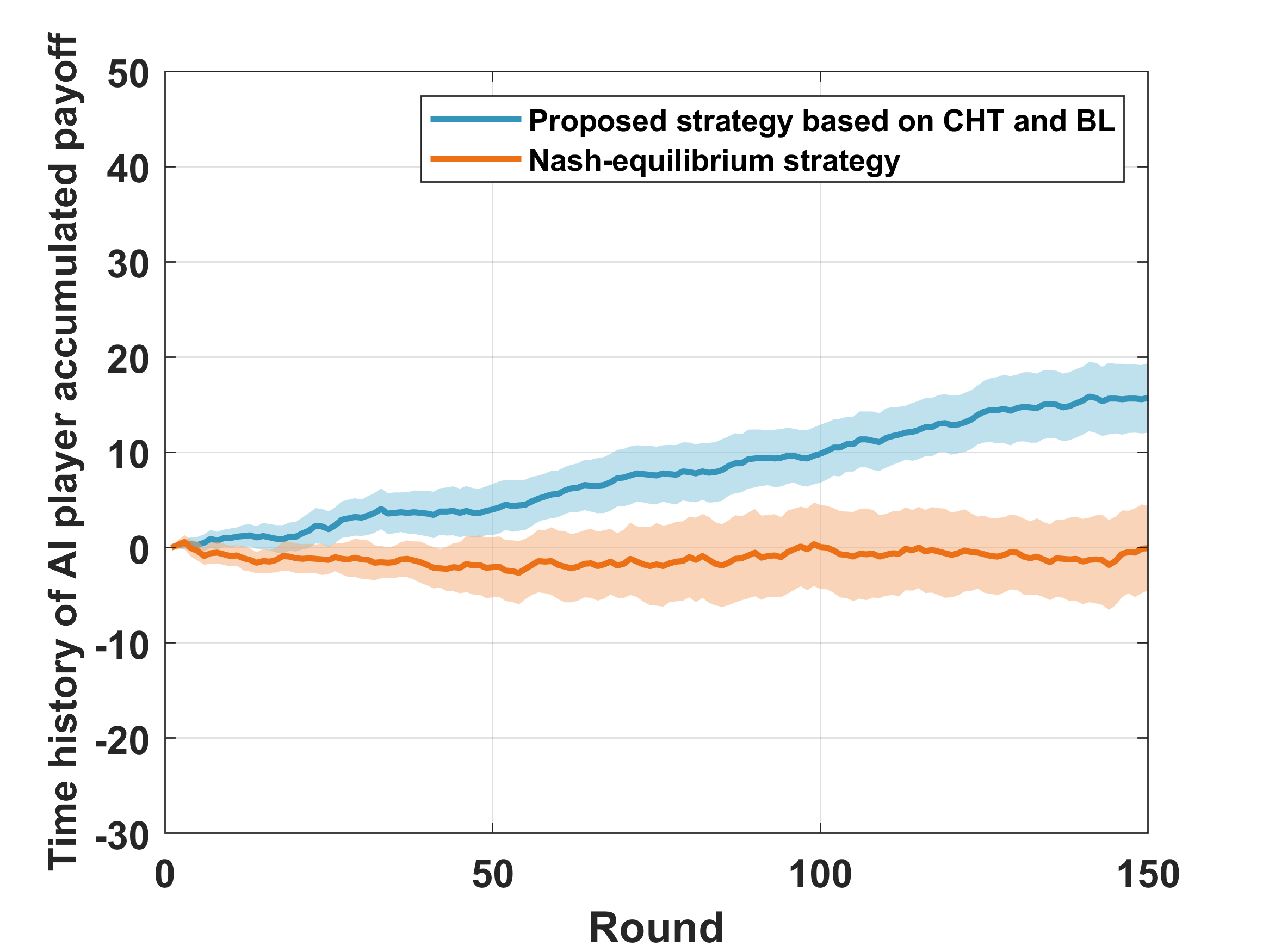,width = 0.85 \linewidth, trim=0.0cm 0.0cm 0cm 0cm,clip}}  
\end{picture}
\end{center}
      \caption{The evolution of accumulated payoff of the AI player against human players.}
      \label{fig: human_result_1}
      \vspace{-0.1in}
\end{figure}

We have collected data of $30$ human participants. We plot the evolution of accumulated payoff of the AI player as the game progresses, $\sum_{k=1}^{t} r^2_{k} = - \sum_{k=1}^{t} r^1_{k}$, in Fig.~\ref{fig: human_result_1}. The thick blue line represents the mean and the light blue shaded area represents the $95\%$ confidence tube of the data for the game-run with our proposed strategy. The thick orange line represents the mean and the light orange shaded area represents the $95\%$ confidence tube of the data for the game-runs with the Nash-equilibrium strategy. It can be observed that when the AI uses our proposed strategy, its accumulated payoff keeps increasing as the game progresses. In contrast, when the AI uses the Nash-equilibrium strategy, its accumulated payoff remains close to $0$. This observation verifies the fact that as long as one of the two players applies such a Nash-equilibrium strategy, i.e., chooses between left or right with equal probability in each round, the game will end in a draw in expectation.

\begin{figure}[ht!]
\begin{center}
\begin{picture}(200, 140)
\put(  0,  -6){\epsfig{file=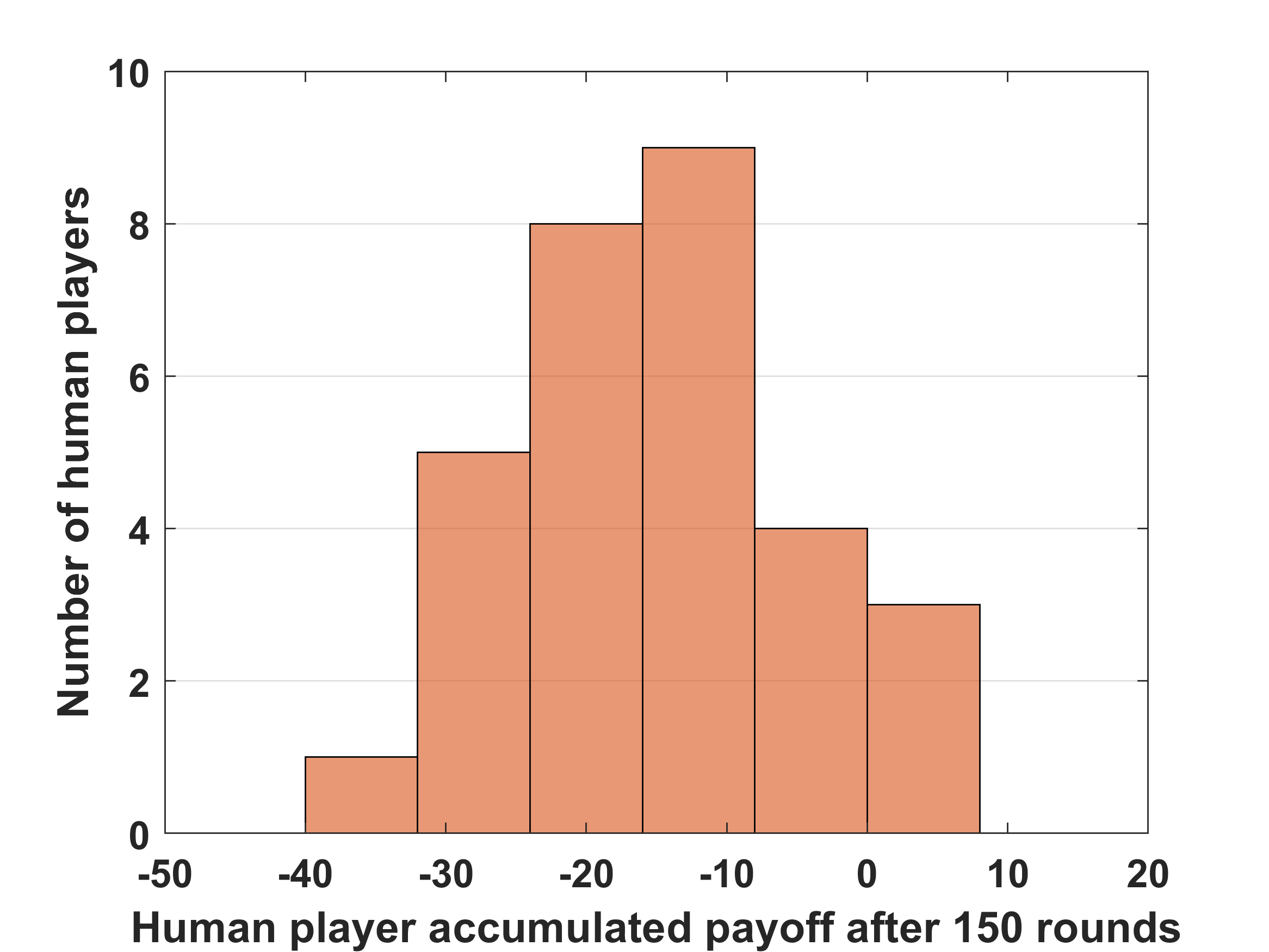,width = 0.85 \linewidth, trim=0.0cm 0.0cm 0cm 0cm,clip}}  
\end{picture}
\end{center}
      \caption{The histogram of accumulated payoffs of human players against the AI player using our proposed strategy.}
      \label{fig: human_result_2}
      \vspace{-0.1in}
\end{figure}

Fig.~\ref{fig: human_result_2} shows the histogram of accumulated payoffs after $150$ rounds, $\sum_{t=1}^{150} r^1_{t}$, of the $30$ human participants corresponding to their game-runs where the AI uses our proposed strategy. It can be observed that the AI using our proposed strategy beats $90\%$ of the human players.

Finally, we are interested in the results when player~1 can be perfectly modeled by our proposed human player's model incorporating level-k reasoning and probabilistic reasoning level transitions. Therefore, we create a ``fake human'' by letting her make decisions according to the ``softmax level-$k$ decision rules'' defined by \eqref{equ:level_k} and Table~\ref{tab:level_k} in each round, with her reasoning level $\sigma_t$ probabilistically transitioned according to the transition model defined by \eqref{equ:level_transition_2} and \eqref{equ:indistinguishable} between consecutive rounds, i.e., a model that perfectly satisfies all of our assumptions. In particular, we randomly generate the values for $q_1^+$, $q_2^+$, $q_1^-$, and $q_2^-$ according to uniform distributions on $[0,1]$.

\begin{figure}[ht!]
\begin{center}
\begin{picture}(200, 140)
\put(  0,  -10){\epsfig{file=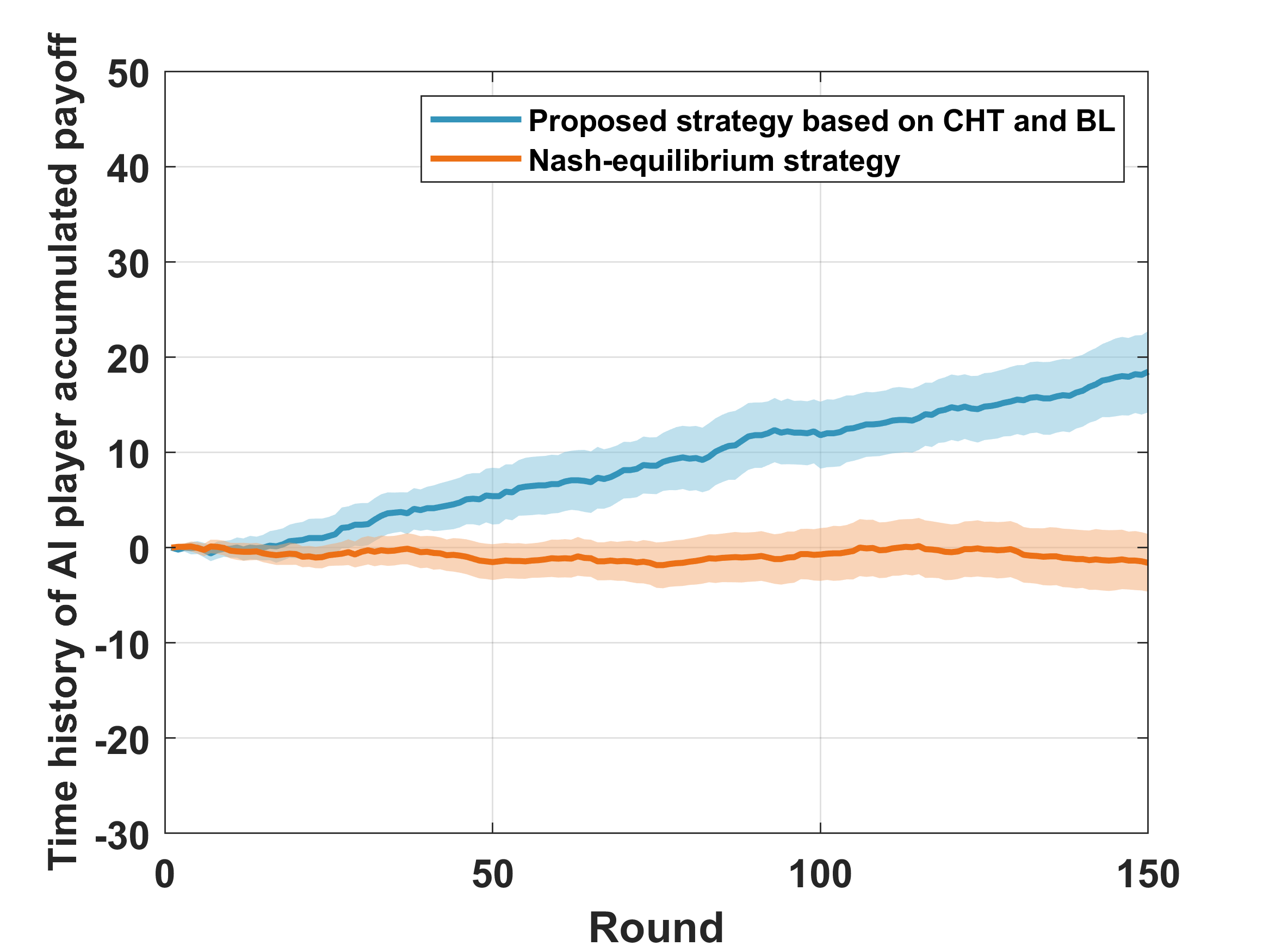,width = 0.85 \linewidth, trim=0.0cm 0.0cm 0cm 0cm,clip}}  
\end{picture}
\end{center}
      \caption{The evolution of accumulated payoff of the AI player against ``fake human'' players.}
      \label{fig: fake_human_result_1}
      \vspace{-0.12in}
\end{figure}

\begin{figure}[ht!]
\begin{center}
\begin{picture}(200, 140)
\put(  0,  -8){\epsfig{file=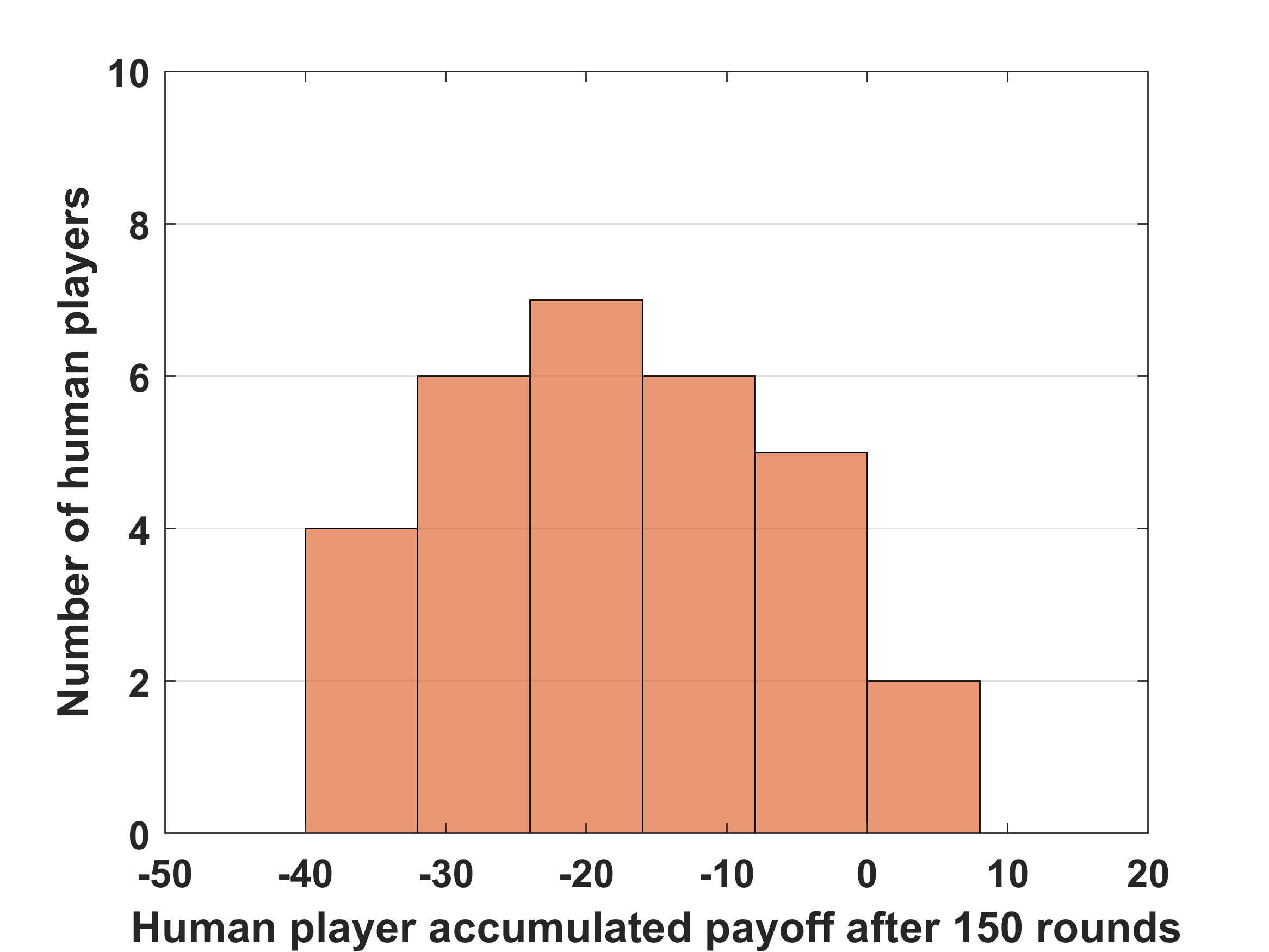,width = 0.85 \linewidth, trim=0.0cm 0.0cm 0cm 0cm,clip}}  
\end{picture}
\end{center}
      \caption{The histogram of accumulated payoffs of ``fake human'' players against the AI player using our proposed strategy.}
      \label{fig: fake_human_result_2}
      \vspace{-0.1in}
\end{figure}

We plot in Figs.~\ref{fig: fake_human_result_1} and \ref{fig: fake_human_result_2} the same results as the ones of Figs.~\ref{fig: human_result_1} and \ref{fig: human_result_2} but with the data of real human players replaced by the data generated by ``fake human'' players. We remark that the $q_1^+$, $q_2^+$, $q_1^-$, $q_2^-$ values are regenerated for each new game-run, thus their values may be different for different 
game-runs, representing the fact that different humans may have different transition models. Furthermore, their true values are unknown by the AI, and the AI has to estimate their values during the game.

It can be observed that the results of AI against real human players in Figs.~\ref{fig: human_result_1} and \ref{fig: human_result_2} are close to those of AI against ``fake human'' players in Figs.~\ref{fig: fake_human_result_1} and \ref{fig: fake_human_result_2}, although the growth of accumulated payoff of the AI player against real human players is slightly slower than that against ``fake human'' players. Note that the latter corresponds to the ideal case where the ``human'' player's behavior perfectly matches the model prediction. Nevertheless, the similarity of the results implies, indirectly, that our proposed human player's model may have captured some crucial features in human decision making in the game. And in the light of this observation, it is reasonable to envision that the proposed approach to modeling human behavior in interactive and sequential decision-making scenarios can find its utility in a broader range of applications involving human-machine interactions.

\section{Summary}

By leveraging cognitive hierarchy theory and Bayesian learning, our AI algorithm beat most human players in a repeated penny-matching game. Our approach to modeling human behavior in the game may be extended and used in other applications involving human-machine interactions.

\bibliographystyle{IEEEtran}

\balance
\bibliography{Ref.bib}
\end{document}